# Large Language Models for Supply Chain Decisions[1]

David Simchi-Levi (MIT), Konstantina Mellou (Microsoft Research), Ishai Menache (Microsoft Research) and Jeevan Pathuri (Microsoft)

**Abstract:** Supply Chain Management requires addressing a variety of complex decision-making challenges, from sourcing strategies to planning and execution. Over the last few decades, advances in computation and information technologies have enabled the transition from manual, intuition and experience-based decision-making, into more automated and data-driven decisions using a variety of tools that apply optimization techniques. These techniques use mathematical methods to improve decision-making.

Unfortunately, business planners and executives still need to spend considerable time and effort to (i) understand and explain the recommendations coming out of these technologies; (ii) analyze various scenarios and answer what-if questions; and (iii) update the mathematical models used in these tools to reflect current business environments. Addressing these challenges requires involving data science teams and/or the technology providers to explain results or make the necessary changes in the technology and hence significantly slows down decision making.

Motivated by the recent advances in Large Language Models (LLMs), we report how this disruptive technology can *democratize supply chain technology* – namely, facilitate the *understanding* of tools' outcomes, as well as the *interaction* with supply chain tools without human-in-the-loop. Specifically, we report how we apply LLMs to address the three challenges described above, thus substantially reducing the time to decision from days and weeks to minutes and hours as well as dramatically increasing planners' and executives' productivity and impact.

## 1. Motivation

Modern supply chains are complex, containing multiple tiers of suppliers, customers, and service providers. Optimization and a variety of related data science tools have been widely utilized for decision making in supply chains including, for example, tools for supply chain planning, network design, procurement and supply chain resiliency. These tools have been applied across multiple industries to allow supply chain planners and executives to make decisions that provide efficiency gains, substantial cost reductions and better customer service.

However, many of these decision-making processes require planners to understand and explain certain decisions recommended by the tools, analyze multiple scenarios, provide what-if analysis, or quickly update the tools to reflect current and changing business environment. Unfortunately, planners are typically not equipped with the necessary background in data science and optimization, resulting in time-consuming back-and-forth interactions with information technologists, data scientists and engineers.

---

[1] Forthcoming chapter in *AI in Supply Chains: Perspectives from Global Thought Leaders*, edited by Maxime C. Cohen and Tinglong Dai, and part of the *Springer Series in Supply Chain Management* (edited by Prof. Chris Tang). The chapter expands and provides more technical details on the concepts and framework described in our recent Harvard Business Review article [3].

Large language models (LLMs) have recently emerged as a promising tool for assisting humans with a wide variety of tasks, such as writing documents, presenting work, coding and health diagnosis. Generative multimodal LLMs, such as Meta's Llama, Google's Palm or Gemini and OpenAI's GPT, are being rapidly integrated within popular software services, such as Microsoft Office or Google Docs and Sheets, for answering questions and increasing productivity through simple, language-based interactions with an underlying information technology. In all these cases, the LLM technology uses structured (numeric) and unstructured (text, image) data to assist and provide recommendations.

For example, in procurements, LLMs have been applied to automate spend analysis, monitor contract compliance, and create customized templates for a variety of contract types. In revenue management, to give another example, LLMs have been applied to leverage historical contracts, win and loss data, and customer characteristics to recommend the price for a new deal. Finally, using detailed information about products and services, LLMs have been applied to recommend a customized configuration package in a business-to-business setting, driven by customer characteristics and market behavior.

In this article, we report how state-of-the-art LLMs can be applied for reasoning about supply chain data and tools used for Sales and Operations Planning (S&OP), Network Strategy, Supply Chain Resiliency and Inventory Optimization. At a high level, the goal of applying LLM technology for supply chains is to *automate key decision-making processes*, without compromising the quality of the decisions.

After providing a brief overview on LLMs (Section 2), we describe three directions in which we have applied this emerging technology: providing insights from data (Section 3), answering what-if questions (Section 4), and updating the supply chain tools to represent current business environments (Section 5). We describe in Section 6 how the LLM-based technology has been used to enhance productivity and efficiency for the supply chain of a public cloud provider (Microsoft) and provide additional examples from procurement (Section 7). We conclude by discussing existing challenges and future opportunities.

## 2. On Large Language Models

A large language model (LLM) is a machine learning model that is trained on extensive data and can be applied for a variety of use cases. In the *training* phase, an LLM learns statistical patterns, word relationships, and contextual information from diverse sources, such as books, articles, websites, and code repositories. At a high level, an LLM is trained to predict the next word following a user sentence (also refers to as input, prompt, or query). This is done by converting every word in the sentence into a sequence of numbers that include the identity of the word, its position in the sentence, and its relationship to other words in the same sentence. It is not difficult to realize that given the ability to predict the next word, LLMs can therefore generate a complete sentence, paragraph or an entire story that has the feel and look of a human-written text.

As mentioned above, an LLM is not limited to a specific language or domain and can be used for a variety of tasks. After the LLM is trained, it can be used in real time to generate output for chatbots, translation, writing assistance, coding, poem and story composition – even for patterns and tasks that it has not explicitly seen before. The output generation process is commonly referred to as the *inference* phase.

Both the training and inference phases are carried out on top of Graphics Processing Units (GPUs), specialized hardware that accelerates the underlying mathematical computations.

Different strategies have been employed to adapt LLMs for a specific application. The most common approaches are fine-tuning and in-context learning. Fine-tuning involves tweaking the parameters of the LLM through training with domain-specific data, which results in a new model specialized for a specific domain. In-context learning is an alternative approach, which involves incorporating a few training examples into the prompt (or query) without changing the LLM. The idea here is to feed the existing LLM with a domain specific list of (prompts, answers) pairs and have the LLM learn from this list of examples. A key advantage of this approach is that the LLM can be used as-is, without requiring any parameter tuning.

Next, we describe how LLMs are integrated with supply chain planning tools to help address three use cases: (i) Understand decisions; (ii) Answer what-if questions; and (iii) Engage in an interactive planning process. We provide a figure for these use cases, see Section 4, that describes the interaction between the user, the LLM, the supply chain planning tool and the company's internal data.

## 3. Data Discovery and Insights

The most immediate use of LLMs is in assisting business operators understand the current situation of the supply chain by facilitating data discovery and providing insights. Consider a classic supply chain with a certain number of suppliers of raw material, factories for producing certain products and retailers that sell these products. With LLMs, planners can ask in plain language questions such as "How much raw material of type T does supplier S have today?", or "What is the cheapest shipping option from factory F to retailer R?". An LLM can translate these questions into data science queries which are in turn fed into the company's data repository (e.g., a SQL database) for obtaining the numerical result. The result can easily be brought back to the planner in a complete sentence form, again by standard use of LLMs.

Observe that in the above-described data flow, the LLM can be utilized as a cloud service, whereas the propriety data does not need to be transferred to the LLM. This is an important advantage from a privacy perspective, as companies may be reluctant to send their data to a third-party LLM service.

A straightforward extension of the above concept is using the LLM not only for understanding the current state of the supply chain, but also for explaining supply chain decisions made by the supply chain tool and providing insights. For example, the planner may ask "How many products of type T are being shipped today to retailer R?". An operator can ask questions not only about the current plan but also about historical trends. For example, "Which was the most productive factory last week?", or "Please report the fraction of instances where the total shipping cost exceeded 50,000 dollars last month". Answering such questions involves accessing the supply chain's plan, which is typically derived by a supply chain planning tool and then hosted in the company's data repository.

## 4. Answering What-If Questions

To get a full picture of the underlying decision processes, operators may require more than just explaining the current supply chain plan. In this section we describe an additional level of sophistication: using LLMs to answer what-if questions. We will discuss the technical innovation that enables answering such questions and provide some examples.

Returning to our generic supply chain example, operators may ask questions such as: "What would be the additional cost if the overall product demand increases by 15%?", "What would be the additional cost if retailer R can use products only from factory F?", "Can we still fulfill all demand if we shut down factory F?", "What would be the cost reduction if raw material of type T is $1 cheaper, per unit?", "Given new tariff levels, how should we reroute, or where should we produce, products for the United States and what is the impact on lead time, service level and inventory?" .

How can an LLM answer all these types of questions accurately and efficiently? One important aspect that enables using LLMs is that supply chain planning tools use mathematical models, typically optimization methods, to generate a recommendation (i.e., a supply chain plan) and these optimization methods have their own "language". Indeed, many optimization tasks are written in the form of *mathematical programs*, which transfer the structure of the supply chain and all the business requirements into a mathematical model. A key idea behind our design is not to replace the mathematical model with LLMs but rather use the mathematical model in tandem with LLMs.

More specifically, the LLM is responsible for translating the human query into a "mathematical code". Due to the nature of the what-if questions, this code can be regarded as small change to the original mathematical model that was used to obtain the plan. For example, forcing a retailer to use products from a particular factory can be done by adding a mathematical requirement (referred to as constraint) that prohibits other factories from sending products to that retailer. This small change in the mathematical model is then fed to the supply chain tool to produce a modified plan, which is used only for comparison with the existing one. As before, the output of the (new) mathematical model can be then passed through the LLM for producing the answer in human language (see the figure below).

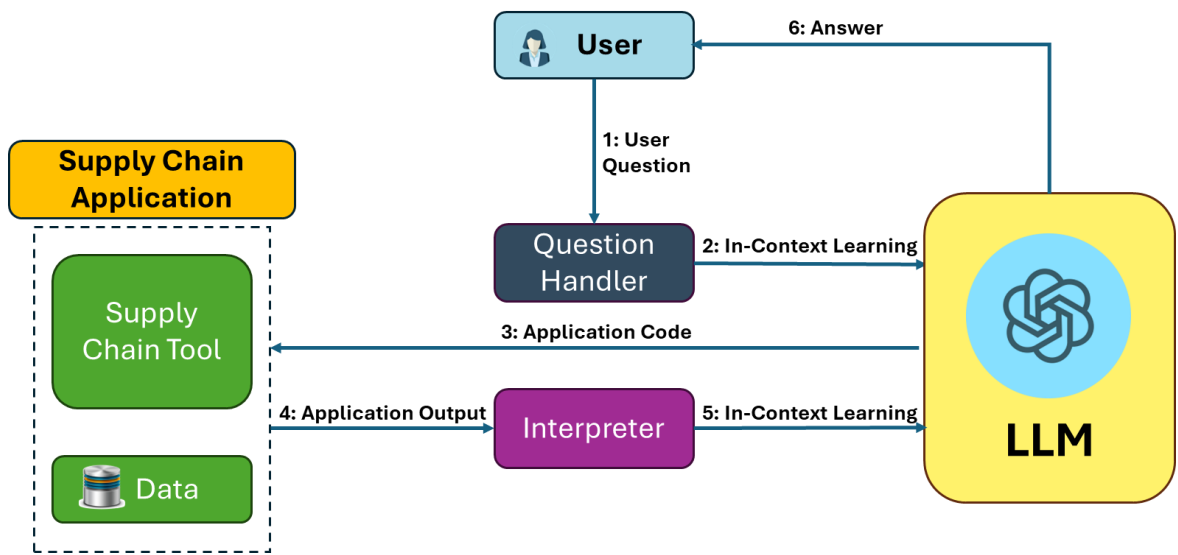

*Figure 1.* *Data flow of LLM-based technology for supply chains. Upon receiving a user question, the question handler processes it, appends examples, and sends the resulting query to the LLM. The LLM outputs application code, which*

*may invoke a supply chain tool (e.g., optimization solver) and/or utilize the application's data repository. The output of the application is then passed to the interpreter, which translates the application's output to plain language. Note that the application data need not be passed to the LLM.*

An important question in this general data flow is how we can get the LLM to produce the necessary mathematical code. The basic idea is to generate a large repository of questions and code answers. Some of these question-answer pairs are then appended to the user query and are used as examples as part of the in-context learning approach described earlier; see [1] for more details.

## 5. Interactive Planning

Going beyond answering what-if questions, we observe that LLM technology can be used more broadly for *interactive* planning, where planners would like to update the model to reflect the current business environment or the LLM provides updates to the planner on a change in business conditions.

For example, a planner may receive real-time information that a specific manufacturing facility is down for seven days due to a winter storm in its area. The planner would like to update the sales and operation plan (S&OP) to account for the disruption. Specifically, the planners would have to engage the IT and data science team to make the necessary plan adjustments, a process that might be time consuming. With LLM-based technology, the planner can directly ask the technology to generate a new plan that avoids using the disrupted manufacturing facility. Of course, it is possible that the new plan will not be able to satisfy all forecasted demand and LLM-based technology will report back, using the existing supply chain planning tool, not only the (updated) plan, but also the new supply chain cost, demand lost, and the impact on profitability.

The need to change the supply plan may also be driven by LLM-based technology. For example, analyzing historical shipment data from a specific supplier, LLM-based technology may generate an alarm that the lead time from the supplier has increased significantly over the last few months. Based on that, LLM-based technology will predict the time when the next shipment is likely to arrive and will communicate that to the planner. Recognizing that the increase in delivery lead time will adversely affect service level, unless corrective actions are taken, the planner may ask the LLM-based technology to re-run the planning tool with the new lead time information. The new plan, generated by the planning tool and translated to the planner, may involve expedited shipments or transfer of inventory.

As a more advanced direction which we have only started exploring, we envision that LLM-based technology would support *end-to-end decision-making* scenarios. For example, users may be able to describe in plain human language the decision-making problem that they wish to solve. LLM-based technology in turn will either generate a mathematical model corresponding to the user description, or directly provide the decisions via internal logic. The challenge in this case is that we do not yet have the appropriate tools to validate a complete mathematical model generated by the LLM-based technology. Indeed, our experience so far is that the technology is currently capable of modeling simple supply chain structures but is not capable today of reliably modeling a more complex, industrial, environment. We still need tools that will either validate the model generated by LLM-based technology or identify what is missing and how to correct the initial model generated by the technology.

# 6. Production Deployment – Microsoft's Cloud Supply Chain

## Background

Cloud computing is a multi-billion-dollar business that draws substantial capital investments from large companies such as Amazon, Microsoft and Google. Large cloud providers need to accommodate the growing demand for computing resources while avoiding over-provisioning of hardware and operational costs. Towards that end, cloud providers periodically make hardware deployment decisions, taking into account many cost considerations (e.g., shipping, depreciation of hardware) and operational constraints (compatibility, capacity, inventory, operational throughputs, etc.). Azure, the Microsoft cloud, consists of tens of regions worldwide and hundreds of datacenters. Demand is disclosed by internal business groups that own different cloud offerings and services (e.g., Azure Storage, Azure Virtual Machine and Office'365 in Microsoft). The hardware unit used for demand requests is a cluster, a set of servers jointly installed in the data center.

At a high level, the demand is exposed through a *demand plan*, which summarizes important information about demand requests for the next year or so. The information includes the business group, the type of hardware required, the region where the hardware should be docked and the ideal dock date. Given the demand plan, Microsoft periodically runs sophisticated optimization solvers to produce a *fulfillment plan,* namely assigning and shipping actual hardware to the datacenters. Microsoft employs *planners* to oversee the demand and fulfillment plan. These are professionals that have the business context but are not necessarily data science or optimization experts. The planners' tasks include understanding and adjusting the demand plan details, confirming that the fulfillment plan meets business needs, and ensuring the execution of the underlying decisions within the fulfillment plan. We describe below two examples on how Microsoft has used LLM-based technologies to facilitate the planners' job in face of the increased scale and complexity of cloud supply chain operations. While the examples are taken from Microsoft's experience, the methodology and insights are relevant more broadly -- not only for other public clouds, but also for supply chain management at large.

## Example 1 – Demand Drift Analysis

While hardware requirements are requested separately by the different business units, a single demand plan is generated regularly by the supply chain team, serving as an important aggregate for supporting the customer requirements. Planners and executives monitor the changes in the demand plan (i.e., the *demand drift*) monthly to ensure that the new plan continues to meet the customer requirement while being commensurate with budget guidelines. The task of evaluating the changes to the demand is traditionally done manually by the planners who often involve program managers, data scientists and engineers from different business units in a process that takes several days. Once the changes are understood, an executive summary is prepared by the planners to explain the changes to the demand for each region.

Microsoft has applied LLM-based technology to ingest the demand plan from different time periods and automatically generate an email report with all the changes to the plan. The report details the person making the change and the input that led to the change. The email also points out potential errors that planners can review and adjust. With this solution, planners can complete the demand drift analysis without involving other parties, i.e., without requesting the help of data scientists.

To illustrate a specific scenario, suppose that demand in the new plan is lower than the old plan (e.g., demand measured in total server count). The email can point to the exact reason why the demand decreased. One reason could be the introduction of a new and more efficient generation of hardware, which requires fewer servers. Another possible reason could be that certain customers have reduced their requirements. The LLM automatically identifies the reason, which is incorporated into the e-mail.

### Example 2 – Understanding Fulfillment Decisions

Microsoft uses optimization tools to derive the fulfillment plan which focuses on assigning and shipping servers from warehouses to data centers. For each demand request, the main decisions consist of (i) the hardware supplier that will be used to fulfill the demand, (ii) the timeline of the deployment - in particular, the cluster's dock-date, and (iii) the particular cluster's deployment location in the data center (selection of a row of tiles to place the cluster on). The goal is to minimize the total cost that consists of multiple components, such as shipping costs and delay cost of the clusters compared to their ideal dock-date, while respecting a multitude of constraints. See [2] for more background.

When planners receive the outcome of an optimization tool, they can confirm that it meets business needs and ensure the execution of the decisions is completed as planned. However, the increased complexity of the underlying optimization problems prevents immediate clarity for the reasoning behind each decision. Consequently, planners often reach out to the engineers and data scientists that developed the optimization tools for obtaining additional insights. Oftentimes, planners and engineers have multiple rounds of interaction around understanding an issue or exploring what-if scenarios, which might result in a delay of days until obtaining a satisfactory solution. Microsoft has connected the optimization tools to the LLM-based technology for answering what-if questions (see Section 4). Planners are now able to automatically obtain answers to questions such as "What is the cost increase if we dock a certain demand a week earlier?", "How will cost change if we disactivate a certain warehouse for one week?".

### Evaluation and Deployment

The LLM-based technology was tested extensively in research and product teams before it was shipped in production. The scenarios tested were not limited only to Examples 1-2 described above, but included synthetic examples from other domains, such as a manufacturing scenario with raw-material suppliers, factories and retailers.

A general evaluation methodology has been developed to test the effectiveness of the LLM technology. In a nutshell, the methodology involves creating a set of test questions for each example. Some of the questions are challenging, for example include incorrect grammar, or atypical questions for the specific domain. Multiple language models and training approaches have been tested, including in-context learning for the larger models such as GPT-4 (OpenAI), and fine-tuning for smaller open-sourced models such as Llama-2 from Meta and Phi-2 from Microsoft.

The current deployment uses GPT-4 and achieves around 90% accuracy. Nonetheless, the research team is actively examining different open-source models, and it seems possible that smaller language models may be used in the future (see Section 8 for more details). Microsoft is applying a gradual deployment strategy where the technology currently supports the most common what-if questions. The idea here is to monitor user interactions, accuracy, and fallback mechanisms, and expand coverage over time.

### User Experience and Impact

Using LLM-based technology presents a substantial change in the working practices of planners. To enable an efficient adoption of the new technology, the cloud supply chain organization introduced an online portal with a variety of useful links and online courses on AI and LLMs. The deployment of LLM-based technology came with a careful design of a user-friendly interface. For example, planners can clearly distinguish between the demand drift analyzer (Example 1) and the fulfillment plan's Q&A system (Example 2) and choose the one that is relevant for them. For the Q&A system, planners are provided with the types of questions that the tool currently supports, to avoid redundant interactions.

Using LLM-based technology has substantially increased the planners' productivity and requires less personnel to resolve an issue. For the drift analysis, it is taking planners only minutes to extract all the changes that were made to the new demand plan, including the root cause of the change. Before using LLMs, such analysis would take planners about a week. Similarly, the fulfillment plan Q&A system enables answering key questions efficiently, eliminating the time-consuming manual processes. The net outcome is response times of few minutes instead of days without compromising on the accuracy of the answer. The system is estimated to have saved planners around 23% of fulfillment investigation time. More recently, we have integrated insight and recommendation modules into our fulfillment systems (e.g., add a new transit lane between two locations). The insights and recommendations are generated by the LLM-based technology (see Section 5).

We conclude this section with some quotes from planners that attest to their boost in productivity: "I am excited to see the AI-powered assistant answering questions that I would normally have to manually investigate. We had to file several tickets to the engineering team in the past to deep dive the root cause, cost, and all different kinds of issues. With the AI-powered assistant, it saved both users and engineering team a good amount of time on the manual process"; "Our system uses complex logic with tens of interdependencies, and understanding the output of the fulfillment plan is challenging. Applying the power of AI tailored to our business challenges radically changes my daily way of work and empowers me to achieve more."

## 7. Procurement Examples

Beyond the implementation of LLM-based technology for supply chain decisions at Microsoft, in this section we report implementations to improve the management and execution of procurement decisions.

In the chemical industry, supply contracts include not only pricing, quality, lead time, and risk mitigation information but also detailed regulatory requirements. A manufacturer typically has thousands of different contracts that require manual and time-consuming processes to identify contract compliance with all business and legal requirements. Recently, a large manufacturer has applied LLMs to re-design contract templates that unify contract details by product category. It allows the manufacturer to streamline contracts and quickly provide an initial draft that includes all legal and business requirements relevant to a new deal.

In the automotive industry, OEM such as Ford, Toyota or GM, have thousands of suppliers and multiple contracts with each supplier. These contracts specify the details of the price paid by the OEM, quality requirements, lead times, and the resiliency measures suppliers need to take to ensure supply. Feeding the LLMs with data of thousands of contracts, one OEM was recently able to identify price reductions specified in the contracts but not followed by procurement. Specifically, some of the contracts provide the OEM with price discount if the OEM's order quantities are larger than a certain threshold. Because of contract complexity, procurement typically did not follow on these incentives, until the LLMs referred decision makers to the opportunity. The result was millions of dollars in procurement savings.

In these procurement examples, the LLM-based technology is used for providing important insights from data (see Section 3) by *direct* interaction with the company's historical data (Excel and Word documents, PDF files, or ERP and CRM technologies), without involving additional supply chain tools. The examples illustrate the potential of LLM-based technology to add significant value in procurement processes, both by simplifying and unifying contract design, generating new contracts and making sure that procurement personnel follow on contract details during the life of the contract. This leads to significant time and cost savings.

## 8. Discussion

LLMs is a disruptive technology that will shape supply chain management in years to come. The examples reported in this paper demonstrate a substantial increase in productivity and planner satisfaction. We conclude this chapter by discussing how companies can get started with incorporating LLM based technology for supply chain management and highlight some of the challenges for broader applicability.

One general piece of advice for getting started with LLM-based technology is to ensure that the organization has clean, well-structured, and accessible data. Newcomers should begin with tasks that are mundane and repetitive yet offer high impact with low risk and low complexity. For example, problems related to data discovery and insights (see Sec. 3)—such as generating document summaries, reducing manual copy-paste between tools, or handling common customer inquiries—are ideal starting points. Answering "what-if" questions that involve invoking optimization tools is somewhat more involved, as it requires the optimization tools to be mature and reliable, with well-defined interfaces (e.g., APIs) that LLMs can easily call based on user input.

While this chapter primarily draws on examples from large enterprises, we believe that small companies and businesses can equally benefit from the methodology described. A key enabler is that LLMs are typically accessed as a service—either directly through LLM providers or via major cloud platforms, eliminating the need for companies to invest in costly GPU hardware. Moreover, the software ecosystem for incorporating LLMs has significantly matured, with many high-quality open-source tools now available, creating a convenient and accessible entry point for smaller organizations.

Although we expect the adoption levels to increase over time, we identify a collection of challenges that require careful attention before LLMs can become pervasive across industries.

**Learning to ask questions.** Optimization is a very precise "language". Human interaction, especially among people that are not optimization experts, may be less formal and even ambiguous from a decision-making perspective. For example, a user may ask "Can we utilize factory F better?". "Better" can have multiple interpretations – lower costs, higher throughput, a more balanced throughput throughout time, etc. Each interpretation would lead to different decisions. Consequently, users may need to be trained to ask more precise questions, or, in turn, the LLM framework would have to generate clarifying questions, until obtaining an unequivocal question from the user.

**Verifying correctness.** We identify multiple challenges in verifying the output generated by LLMs, especially when the LLM is required to produce substantial parts of code. For example, suppose we let the LLM generate an entire mathematical program from scratch -- how would the system verify that it is correct? Going beyond correctness, how do we ensure that the proposed formulation can be solved in reasonable time? Even with proper verification, we expect LLMs to occasionally produce wrong outputs. Thus, system designers need to dedicate attention to identifying mistakes and recovering from them. See our recent HBR article [3] for further discussion.

**Data availability.** Regardless of the mode of interaction with the LLM, decision making requires clean and reliable data. Organizations will have to put emphasis on collecting, gathering, and cleaning of data. As discussed earlier, system designers aim at keeping proprietary data separate from the LLM-powered framework. Nonetheless, this might not be fully possible in some cases, for example, when user-specific data needs to be added to the query to provide more context to the LLM. Organizations will have to find the right balance between protecting data and providing LLMs with adequate information to execute decision making tasks.

Similarly, data quality challenges may render supply chain planning tools obsolete. For example, missing data is a common problem faced by IT and data scientists. We already have tools for identifying inconsistent data and/or missing data. Of course, identifying poor quality data is not enough. We need tools for improving the quality of the data used by the supply chain planning tools. LLM-based technology may provide such a solution. For example, in [4] we describe how one can apply AI and LLM-based technology to complement existing data by generating synthetic data for experimental design.

**Understanding cost and performance tradeoffs.** In relation to the above, organizations may consider multiple options for aligning LLMs to their specific requirements. Fine-tuning is one such option, however the training process itself might be costly. A different alternative is to append queries with domain-specific data: Retrieval-Augmented Generation (RAG) is a popular technique for efficiently doing so. However, a drawback of appending data is that the query becomes longer, making the cost-per-query more expensive. Orthogonal to these, we are witnessing the emergence of Small Language Models (SLMs) which are shown to be able to execute specific tasks with accuracy approaching that of LLMs [5]. Using these smaller language models may lead to cost reductions in the future. We believe that some decision-making tasks can be carried out with such models with proper data collection and tuning techniques. Finally, LLMs with reasoning capabilities have recently been introduced and have shown impressive results in solving complex tasks in math and coding. It will be interesting to examine how these models can accommodate complex, interactive planning tasks.